\documentclass[letterpaper]{article} % DO NOT CHANGE THIS
\usepackage{aaai2026}  % DO NOT CHANGE THIS
\usepackage{times}  % DO NOT CHANGE THIS
\usepackage{helvet}  % DO NOT CHANGE THIS
\usepackage{courier}  % DO NOT CHANGE THIS
\usepackage[hyphens]{url}  % DO NOT CHANGE THIS
\usepackage{graphicx} % DO NOT CHANGE THIS
\usepackage{natbib}  % DO NOT CHANGE THIS AND DO NOT ADD ANY OPTIONS TO IT
\usepackage{caption} % DO NOT CHANGE THIS AND DO NOT ADD ANY OPTIONS TO IT
\usepackage{multirow} % Added to fix undefined \multirow error
\usepackage{algorithm}

\usepackage{amsmath,amssymb,amsfonts}
\usepackage{algorithm}
\usepackage{algpseudocode}

\usepackage{amsmath}
\usepackage[table]{xcolor}
\usepackage{array}
\usepackage{booktabs}
\usepackage{algorithmicx}

\definecolor{lightgray}{gray}{0.95}
\newcolumntype{C}{>{\centering\arraybackslash}p{\textwidth}}
\usepackage{tabularx}
\newcommand{\release}[1]{}
\usepackage{amsmath,amsthm,amssymb}
\usepackage{bm}

\usepackage[table]{xcolor}

\usepackage{newfloat}
\usepackage{listings}
\DeclareCaptionStyle{ruled}{labelfont=normalfont,labelsep=colon,strut=off} % DO NOT CHANGE THIS
\floatstyle{ruled}
\newfloat{listing}{tb}{lst}{}
\floatname{listing}{Listing}
\title{Reinforcement Learning Enhanced Multi-hop Reasoning for Temporal Knowledge Question Answering}
\author{
    Wuzhenghong Wen\textsuperscript{\rm 1}\equalcontrib,
    Chao Xue\textsuperscript{\rm 2}\equalcontrib,
    Su Pan\textsuperscript{\rm 1}\thanks{Corresponding author.},
    Yuwei Sun\textsuperscript{\rm 1},
    Minlong Peng\textsuperscript{\rm 3}
}
\affiliations{
    \textsuperscript{\rm 1}School of Internet of Things, Nanjing University of Posts and Telecommunications\\
    \textsuperscript{\rm 2}School of Software, Beihang University\\
    \textsuperscript{\rm 3}Fudan University\\
    \{2022070804, supan, 2021070706\}@njupt.edu.cn, 
    xuechao@buaa.edu.cn, mlpeng16@fudan.edu.cn
}

\usepackage{bibentry}
\makeatletter
\renewcommand{\section}{%
  \@startsection{section}{1}{\z@}%
  {10pt} % space above
  {4pt}  % space below
  {\centering\bfseries\large}%
}
\makeatother

\begin{document}

\maketitle

\begin{abstract}
Temporal knowledge graph question answering (TKGQA) involves multi-hop reasoning over temporally constrained entity relationships in the knowledge graph to answer a given question. 
However, at each hop, large language models (LLMs) retrieve subgraphs with numerous temporally similar and semantically complex relations, increasing the risk of suboptimal decisions and error propagation. 
To address these challenges, we propose the multi-hop reasoning enhanced (MRE) framework, which enhances both forward and backward reasoning to improve the identification of globally optimal reasoning trajectories.
Specifically, MRE begins with prompt engineering to guide LLM in generating diverse reasoning trajectories for the given question. Valid reasoning trajectories are then selected for supervised fine-tuning, serving as a cold-start strategy. Finally, we introduce Tree-Group Relative Policy Optimization (T-GRPO)—a recursive, tree-structured learning-by-exploration approach. At each hop, exploration establishes strong causal dependencies on the previous hop, while evaluation is informed by multi-path exploration feedback from subsequent hops. Experimental results on two TKGQA benchmarks indicate that the proposed MRE-based model consistently surpasses state-of-the-art (SOTA) approaches in handling complex multi-hop queries. Further analysis highlights improved interpretability and robustness to noisy temporal annotations.
\end{abstract}

\begin{figure}[ht]
\centering
\includegraphics[width=.95\linewidth]{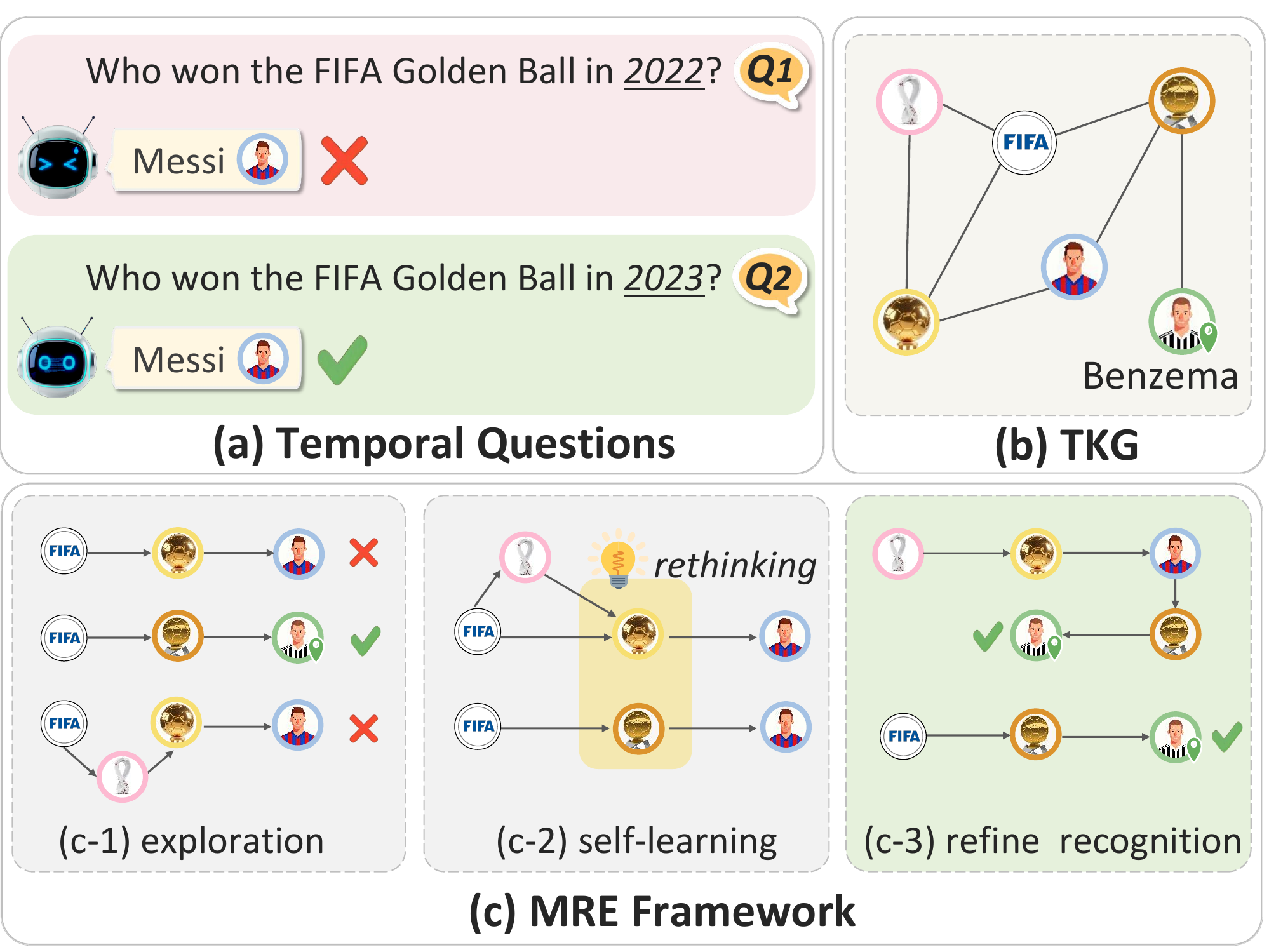}
\caption{Retrieval errors under complex temporal facts (b) and the resolution by the MRE framework (c).}
\label{fig:eurai}
\end{figure}

\section{Introduction}
Temporal Knowledge Graph Question Answering (TKGQA) aims to answer temporally-aware questions by constructing a time-sensitive subgraph centered on a target entity, performing multi-hop reasoning over relevant entities and temporal relations, and ultimately inferring the correct answer. Traditional TKGQA approaches mainly rely on embedding-based methods, which align temporal relational graphs with natural language queries through latent representations~\cite{CHEN2022109134, mhm, MGT}. Although effective for in-distribution queries, these methods often exhibit poor generalization to out-of-distribution or temporally complex scenarios. Leveraging the extensive pre-trained knowledge of large language models (LLMs) and recent advances in long-context modeling~\cite{RoFormer, PI}, there is growing momentum toward applying both commercial~\cite{achiam2023gpt} and open-source~\cite{touvron2023llama, yang2024qwen2.5} LLMs to TKGQA. LLMs show particular strength in multi-hop temporal reasoning, where answering a question requires traversing a sequence of temporally grounded facts~\cite{ARI, qianyihu-etal-2025-time}. This paradigm shift from embedding-based to LLM-based TKGQA opens new avenues for integrating pretrained linguistic knowledge with temporal graph structures, leading to improvements in both reasoning accuracy and robustness.
% can give more  

However, in LLM-based  TKGQA approaches, the complexity of event retrieval ~\cite{dziri, Rek} and ambiguous temporal relations~\cite{timer4, zha2024m} can mislead the language models(LMs), resulting in suboptimal decisions during intermediate reasoning steps. As illustrated in Figure~\ref{fig:eurai}(a), when querying Golden Ball recipients in 2023, the absence of competing candidates enables the LLM to identify the correct answers directly from the subgraphs. In contrast, the year 2022 presents a more challenging case: FIFA issues two distinct Golden Ball awards, and Messi’s World Cup triumph received unprecedented media attention. This prominence misleads the LLM, causing confusion in identifying the correct award recipient (Ballon d'Or). Although selecting Messi as the final answer is incorrect, including him as an intermediate reasoning node reflects a suboptimal but plausible inference. These cases highlight the inherent difficulty in guiding LLMs toward globally optimal trajectories in multi-hop reasoning.

Reinforcement Learning from Human Feedback (RLHF)~\cite{DPO, PPO} is a fine-tuning paradigm that steers models toward globally optimal solutions through preference optimization. 
As an advanced extension of RLHF, group relative policy optimization (GRPO)~\cite{guo2025deepseek}, improves reasoning by comparing multiple candidate outputs and leveraging contrastive learning and enhanced exploration. Compared to traditional methods such as PPO~\cite{PPO}, GRPO offers greater robustness and a stronger capacity for global optimality in reasoning tasks.

Unlike single-turn question answering (QA) tasks, where immediate feedback is available from the final answer, TKGQA involves multi-hop reasoning to retrieve intermediate information before arriving at the final answer. This inherently introduces the problem of reward sparsity~\cite{sparse1, sparse2, sparse3}. As a result, it becomes challenging to guide LLMs to escape local optima during intermediate steps and achieve global optimality throughout the reasoning trajectory, while simultaneously mitigating the effects of sparse rewards in the RLHF optimization process.

% \subsection{Multi-Hop Reasoning Enhanced (MRE) Framework}

To extend GRPO’s global optimization capability from single-turn QA to multi-hop trajectory reasoning, we propose the \textbf{Multi-hop Reasoning Enhanced (MRE)} framework, which comprises three key components:
\textbf{(1) Multi-Trajectory Sampling.} Leveraging GPT-4 with prompt engineering, we sample diverse multi-hop reasoning trajectories from a few-shot dataset under varying temperature settings. Trajectories that produce the correct final answers are identified as positive examples. From these, we construct a fine-grained dataset by extracting each intermediate reasoning step as an independent training instance.
\textbf{(2) Cold Start Supervised Fine-Tuning}. Building on the dataset constructed in (1), we apply supervised fine-tuning to the target model, guiding it to imitate the multi-hop reasoning process and improving its adherence to instructions. 
\textbf{(3) Tree-Group Relative Policy Optimization.} To address the challenge of sparse rewards during exploration, T-GRPO adopts a tree-based search strategy, leveraging the exploration results of constructed subtrees for evaluation and learning. Specifically, a search tree is constructed to perform $g$ rounds of reasoning on the given subgraph, guided by the input question. At each hop, the subgraph expands along different branching directions of subsequent subtrees, and the exploration process continues until the final answer is derived. Upon receiving evaluation signals from the $g$ reasoning trajectories returned by their respective subtrees, group reward is applied to compute their contributions. Finally, the GRPO algorithm is employed to model relative preferences among the explored trajectories within each tree, enabling iterative updates of decision policies back to the root.
Our contributions are summarized as follows:
1) We propose the MRE framework, a LLM-based multi-hop reasoning enhancement framework for TKGQA.
2)We propose T-GRPO, a reinforcement learning method for training LLMs in multi-hop reasoning for TKGQA.
3) Experiments with several TKGQA datasets demonstrate the effectiveness of MRE.

\section{Related Work}
\noindent{\textbf{LLM-based TKGQA}} 
With the development of commercial and open-source LLMs, their application to reasoning tasks in TKGQA is garnering increased attention. \cite{ARI,qianyihu-etal-2025-time,wu2025tablebench,fei2022cqg} propose multi-round interactive prompts for multi-hop reasoning, while\cite{lee-etal-2023-temporal} leveraged in-context learning to enhance open-source model performance. \cite{xia-etal-2024-chain,liang2019adaptive,liu2023time} introduce a hybrid method that combined GNN with LLM voting for multi-hop QA. Despite their strong reasoning capabilities, commercial LLMs struggle to effectively integrate TKG knowledge, limiting their applicability to complex TKGQA tasks. As a result, open-source efforts emphasize specialized TKGQA models. \cite{B2F, xiong-etal-2024-large, chen-etal-2024-self,wang2022dabert,liu2023local} focus on generating high-quality temporal chain-of-thought(CoT) data for fine-tuning, with\cite{B2F, xiong-etal-2024-large,liang2019asynchronous} emphasizing multi-path reasoning and\cite{chen-etal-2024-self} adopting in-context learning. Separately,~\cite{timer4} fine-tuned a rewriter to transform complex temporal constraints into explicit time points, while \cite{rememo,zheng2022robust}, while \cite{rememo,gui2018transferring,dai2025hope} constructs a temporally sensitive pre-training model. However, these approaches mainly aim to optimize single-hop reasoning accuracy, neglecting global optimality across the entire multi-hop reasoning trajectory.

\begin{figure*}[th]
\centering
\includegraphics[width=0.95\textwidth]{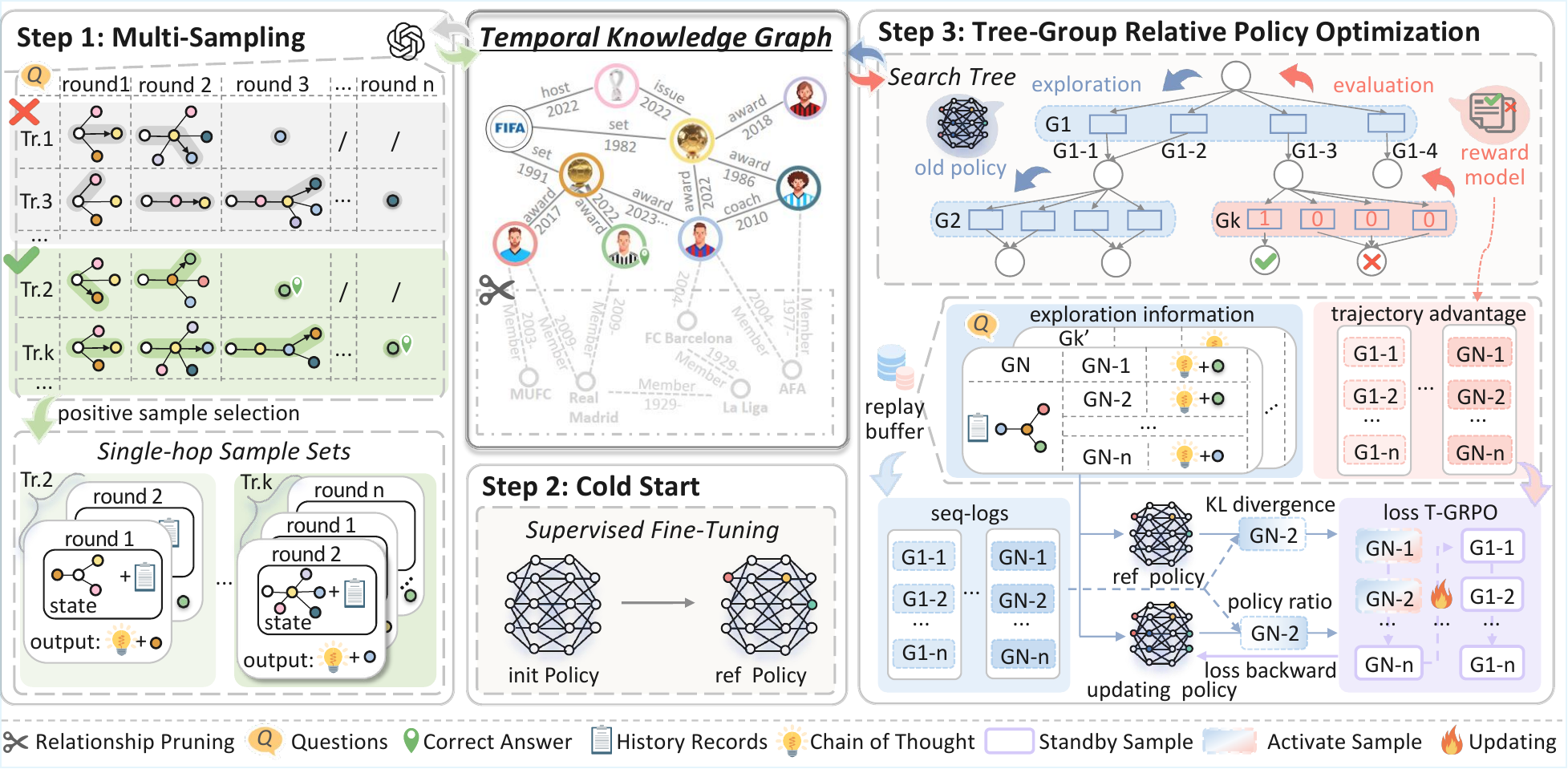}
\caption{Overall architecture of the MRE framework. First, prompt engineering is used to guide the LLM in generating diverse multi-hop reasoning trajectories. Second, valid trajectories are selected for SFT, providing a cold-start policy. Finally, tree-structured exploration with T-GRPO recursively optimizes reasoning paths by leveraging forward decisions and backward feedback across each hops.}
\label{fig:architecture}
\end{figure*}

\noindent{\textbf{RL for Reasoning}}
With CoT based methods~\cite{wei2022chain,yao2023tree,jin-etal-2024-graph,ma2022searching,song2022improving} that enhance LLM reasoning via static supervision, their effectiveness remains constrained by pre-trained knowledge of the model. GRPO~\cite{guo2025deepseek,liu2025structural} addresses this limitation by autonomously discovering CoT trajectories and leveraging high-quality reasoning for self-improvement. To improve training stability and reasoning precision, recent efforts~\cite{yu2025dapo,zheng2025group,xue2024question,wang2025not} focus on enhancing token-level differentiation, allowing the model to assign more informative and fine-grained credit signals across sequences. In parallel, other studies explore rule-guided reasoning by injecting symbolic constraints through RL~\cite{fang2025thinkless,jiang2025think,liu2024resolving,ma2025reasoning,wu2025progressive}. Beyond core reasoning, GRPO is also adapted for downstream tasks~\cite{chen2025rm,wu2025unleashing,liu2025inference}, finance~\cite{liu2025fin,zhu2025dianjin,xue2023dual,li2024comateformer}, and~\cite{lai2025med,li2024local,liu2025stole}. MRE not only harnesses reinforcement learning to enhance the capability of LLMs in achieving global optimality across at each hop, but also introduces a tree-structured exploration and learning strategy, T-GRPO, which effectively mitigates the challenge of sparse reward.

\section{MRE Framework}
In this section, we present a comprehensive description of our proposed MRE framework, which comprises three core components: (1) Multi-trajectory sampling, (2) Cold start supervised fine-tuning and (3) T-GRPO based reinforcement learning. The following subsections will provide detailed explanations of each component's implementation.

\subsection{Task Definition}
Let the temporal knowledge graph (TKG) be denoted as $\mathcal{K} := (\mathcal{E}, \mathcal{R}, \mathcal{T}, \mathcal{F})$, and its 1-hop subgraph  centered on an entity $e$ represented as $\mathcal{K}_e := (\mathcal{E}_e, \mathcal{R}_e, \mathcal{T}_e, \mathcal{F}_e)$. A fact in $\mathcal{K}_e$ can be formalized as ${(e, r, o, \tau) \subset \mathcal{F}_e}$ or ${(s, r, e, \tau)} \subset \mathcal{F}_e$, where $s$, $e$, and $o \in \mathcal{E}_e$ denote the subject entity, the central entity and the object entity, respectively. 
The relation $r \in \mathcal{R}_e$ denotes the relationship between entities, where $\mathcal{R}_e \subset \mathcal{R}_{QA}$ is constrained by the QA task, and $\tau \in \mathcal{T}$ denotes the associated temporal information.
LLM-based TKGQA tasks can be formulated as follows. Given a question $Q$ and an initial entity $e_h$, the LLM first constructs a 1-hop subgraph $\mathcal{K}_{e_{h}}$ centered on the current entity $e_h$. It then performs multi-hop reasoning by selecting a next-hop entity $e'$ from the facts $\mathcal{F}_{e_h}$ or directly producing the answer based on the retrieved information, where the answer can be an entity or a timestamp $\tau$. The optimization objective for multi-hop reasoning in TKGQA is defined as follows:
% $\varepsilon, \beta, \mu$
\vspace*{-0.5cm}

\begin{equation}  \small
	\begin{aligned}  
		& \max_{{\pi}} \mathbb{E}\left[ \sum_{i=0}^{K} \gamma R_i(\mathcal{F}_{e^{i}_h}) \right] \quad \text{s.t.} \quad R_i = 0 \quad \forall i < K
	\end{aligned} 
    \label{opt}
\end{equation}

\subsection{Multi-Trajectory Sampling}
To provide high-quality cold start samples for supervised fine-tuning, we employ GPT-4 to sample multiple reasoning trajectories for each QA under varying reasoning temperature settings . These trajectories are then used to construct a supervised fine-tuning dataset.

We construct multi-hop reasoning trajectories in TKGs by recursively applying 1-hop inference, where each predicted entity becomes one of a new central entity for the next hop. This iterative process proceeds until the final answer is derived. The 1-hop inference and its transition to subsequent hops can be formalized as follows:
\begin{equation}\small
a_k \gets LLM\big(Q, H_k, S_k, E_k\big)
\label{tr.update}
\end{equation}

\begin{equation}
S_{\text{pruning}} = \operatorname*{Top}\text{-}P(Q,\mathcal{F}_E)
\label{eq:purning}
\end{equation}
\noindent

Let $LLM$ denote the sampling model, and let $Q$ denote the input question, and $E_k$ be the  central entity . 
Furthermore, we define $\mathcal{F}_{E_k}$ as the set of textual facts centered on $E_k$.
The action $a_k$ denotes the next-hop decision made by the $LLM$ based on the current subgraph, which consists of the generated CoT and the selection of the next-hop entity $E_{k'}$.
At each round, the retrieved subgraph is constructed by selecting $P$ relevant facts from $\mathcal{F}_{E_k}$ using the relevance retrieval function in Function~\eqref{eq:purning}, where relevance scores are computed based on the similarity between the question $Q$ and each candidate fact $(s, r, o, \tau) \in \mathcal{F}_{E_k}$.
The historical context $H_k$ accumulates the facts traversed in hops and is used to guide the decision-making of $LLM$. 
Following the definition of 1-hop reasoning in TKGQA, we extend it to the multi-hop setting.
\begin{equation}  \small
	\begin{aligned}  
		& Tr_j^t = \{Tr_j^{k} \sim LLM^t(Q,H_{k},S_k,E_k)\}_{k=1}^N
	\end{aligned} 
    \label{tr.tr}
\end{equation}

\noindent Here, $Tr_j^t$ denotes a reasoning trajectory generated by the $LLM$ with temperature $t$, where $N$ is the final step index of the $j$-th trajectory. The set $Tr_{Q_i}$ contains $M$ completed trajectories sampled under different temperature settings for the question $Q_i$, and $a'_j$ denotes the final answer produced by $Tr_j^t$.We then construct a positive trajectory set for $Q_i$ by selecting each trajectory, whose output answer matches the ground-truth $a^*$ as a positive sample. 
\begin{equation}  \small
	\begin{aligned}  
		& Tr_{Q_i} = \{(a'_j,Tr_j^t)\}_{j=1}^M,t\in T
	\end{aligned} 
    \label{tr.q}
\end{equation} 

\begin{equation}  \small
	\begin{aligned}  
		&  Tr_{Q_i}^+ = \{ Tr_j^t \in Tr_{Q_i} \mid a'_j = a^* \}
	\end{aligned} 
    \label{tr.q2}
\end{equation} 
Finally, we build a subset $d$ from the full dataset $D$ to construct a positive trajectory set $Tr_{d \sim D}^+$, 
where 
$Tr_{d \sim D}^+ = \{(Tr_{Q_i}, Q_i, a^*)\}_{i=1}^{|d|}$.

\subsection{Cold start supervised fine-tuning}
To enhance the instruction-following ability of the target LLM in multi-hop TKGQA reasoning and to cultivate core reasoning patterns, we apply supervised fine-tuning based on step-wise decisions extracted from $Tr_{d \sim D}^+$. 
The fine-tuning objective is defined as follows:

\begin{equation}  
\small
	\begin{aligned}  
		& \underset{\theta}{\text{minimize}}  
		& & \sum_{i=1}^{V} \sum_{j=1}^{M} \sum_{k=1}^{N} \text{Loss}\left(y_{ijk}, \pi(Q_i, S_{ijk}, H_{ijk}, E_{ijk}; \theta)\right) 
	\end{aligned}
    \label{fine-tuning}
\end{equation}

In the above objective, $\pi$ denotes the LLM to be optimized, parameterized by trainable parameters ${\theta}$, and $y_{ijk} = (CoT_{ijk}, a_{ijk})$ represents the reasoning process and action at step $k$ of the $j$-th trajectory for the $i$-th question. Here, $V$ represents the total number of questions in $Tr_{d \sim D}^+$, $M$ denotes the number of trajectories associated with a given question $Q$, and $N$ corresponds to the length of the current trajectory $Tr_j^t$. Based on this formulation, we develop a basic multi-hop reasoning model $\pi_{\theta}^{\mathrm{SFT}}$ for TKGQA.

\subsection{T-GRPO} 
As $\pi_{\theta}^{\mathrm{SFT}}$ tends to converge to locally optimal solutions during trajectory reasoning, we introduce T-GRPO to enhance exploration and incorporate the contrastive learning paradigm of GRPO, thereby guiding the model toward globally optimal reasoning paths.
\begin{algorithm}
\caption{Tree-group Searching}
\label{alg:tree_group_search}
\label{alg:searching}
\begin{algorithmic}[1]
\Require $\pi_\theta^{Old}$, $g$, $max\_depth$, $buf\!fer$, $E_{init}$, $S_{init}$, $Q$
\Ensure Sampled buffer $buf\!fer^*$

\State \textbf{function} \text{Searching}($E$, $S$, $H$, $Q$)
\If{the search depth reaches $max\_depth$}
    \State \Return $0$
\EndIf
\State $Group,Reward \gets \emptyset$
\For{$i \gets 1$ \textbf{to} $g$}
    \State $a_i \gets \pi_\theta^{Old}(Q, E, H, S)$ 
    \If{$a_i$ is answer}
        \State compute $Score_i$ based on~\eqref{R_leaf}
        \State go to line (17)
    \EndIf
    \State Extract $E_\text{next}$ from $a_i$
    \State $\mathcal{F}_{E_\text{next}} \gets \mathcal{K}_{E_\text{next}}$ 
    \State $S_\text{next} \gets$ Pruning $\mathcal{F}_{E_\text{next}}$ using (\ref{eq:purning})
    \State $H_\text{next} \gets H \cup S$
    \State $Score_i \gets \text{Searching}(E_{\text{next}}, S_{\text{next}}, H_\text{next}, Q)$
    \State $Group \gets Group \cup \{(a_i, Q, H, S,E)\}$
    \State $Reward \gets Reward \cup \{Score_i\}$
\EndFor
\State Compute $E\_Score$ using~\eqref{R_root}
\State Push $\{Group, Reward\}$ to $buf\!fer$  \Comment{asynchronous}
\State \Return $E\_Score$
\State \textbf{End function}
\State  \text{Searching}($E_{init}$, $S_{init}$, $\emptyset$, $Q$)
\end{algorithmic}
\end{algorithm}

\noindent\textbf{Exploration.} GRPO performs group-wise policy optimization by sampling multiple candidates under the same question and leveraging their relative performance within the group. The sampling process is defined as: $ G \gets \{ a_i \sim \pi_\theta \}_{i=1}^{g}$, where $G$ represents a group of $g$ samples generated by the policy $\pi_\theta$ for a given input.

Multi-hop reasoning in TKGQA exhibits strong causal dependencies, where each decision is conditioned on the inference of the previous hop. Building on this, we define the search tree centered on $E$, denoted as:
\begin{equation}  \small
	\begin{aligned}  
        Tree_{k'} = \{E,\text{Searching}(E,S,H,Q),R_E\} ,\quad k' > 0
	\end{aligned} 
    \label{group}
\end{equation} 

\noindent where $Searching$ function refers to the multi-hop search algorithm~\ref{alg:tree_group_search}, $R_E$ is evaluation of the central entity $E$ after searching and $Tree_{k'}$ is the $k'$-th tree generated during the search process. The core of GRPO is to construct the sampling set $G$ and perform preference optimization iteratively. Therefore, we construct the sampling expression for $G_{k'}$ based on the search tree $Tree_{k'}$:
\begin{equation}  \small
	\begin{aligned}  
    G_{k'} = \{a_{k'}^{(j')} \sim \pi_\theta(Q, H, (S, E)\gets a_{k}^{(j)}) \}_{j'=1}^g, \quad k'> 0
	\end{aligned} 
    \label{group2}
\end{equation} 
\noindent where $G_{k'}$ denotes the group constructed by the policy network $\pi_\theta$ in the $k'$-th tree using the subgraph $S$, which is determined by $E$. Moreover, both $S$ and $E$ are influenced by the information on the $j$-th output in the $k$-th tree, denoted $a_k^{(j)}$, where $a_k^{(j)} = (CoT_{k}^{(j)}, E_\text{next}^{k,j})$.
Consequently, the next-hop subgraph $S_\text{next}^{k,j}$ is constructed accordingly (line 11–13 in Algorithm~\ref{alg:tree_group_search}).

\noindent{\textbf{Evaluation.}}
The sparsity of trajectory-level rewards in multi-hop reasoning poses significant challenges to the convergence of learning algorithms~\cite{sparse2,sparse3}. To address this, T-GRPO builds on the search strategy defined in the previous hop and adopts a tree-structured multi-hop search process. It assigns backward credit by evaluating each node based on the aggregated scores of its downstream search paths. By propagating reward signals along the tree, this approach effectively mitigates the impact of sparse supervision and facilitates more stable and efficient learning.

For a next-hop entity $E_{\text{next}}^{k, j}$ , if action $a_k^{(j)}$ selected in $j$-th inference of the $k$-th group corresponds to the final answer, the associated central entity is denoted $E_{\text{leaf}}^{k, j}$. 
Otherwise, if $a_k^{(j)}$ is the entity chosen for the next hop, we denote it as $E_{\text{root}}^{k, j}$. The evaluation of the leaf entity $ E_{\text{leaf}}^{k, j}$ is conducted by directly comparing it with the ground-truth answer $a^*$, as formalized in function~\eqref{R_leaf}:
\begin{equation}
R_{E_\text{leaf}^{k,j}} = 
\begin{cases} 
1, &  a_k^{(j)} = a^* \\ 
0, & \text{otherwise} \\
\end{cases}
\label{R_leaf}
\end{equation}

The evaluation of non-terminal ${E_\text{root}^{k, j}}$ is based on the reward computed over the group $G^{k,j}_\text{next}$, which is determined by performing a search on the subgraph $S^{k,j}_\text{next}$. %centered on ${E_{root}^{k',j'}}$.
\begin{equation}  \small
	\begin{aligned}  
    R_{E_\text{root}^{k,j}} = \frac{1}{g} \sum_{j'=1}^{g}{R(a_{k'}^{(j')})}
	\end{aligned} 
    \label{R_root}
\end{equation} 

\noindent In Function~\eqref{R_root}, $a_{k'}^{(j')}$ denotes an inference result generated in $G^{k,j}_{\text{next}}$, which produces a total of $g$ responses. The quality of each $a_{k'}^{(j')}$ is assessed according to its corresponding next-hop information.

\paragraph{Storage.}
After evaluating a group of sampled trajectories, we store the resulting group-level sampling data into an asynchronous buffer, decoupling trajectory evaluation from the GRPO training process.
Specifically, the storage structure for the $k$-th tree is defined as:
$G_k = \{ (a_j, Q, H_j, S_j, E_j, R_{E_j}) \}_{j=1}^{g}$ (line 20 in Algorithm~\ref{alg:tree_group_search}). Once a storage structure is fully sampled, it is forwarded to the GRPO algorithm for parameter updates.

% 训练
\noindent{\textbf{Training.}} In the GRPO training framework, three model variants are maintained throughout the optimization process: the update model $\pi_\theta$, the reference model $\pi_{\theta}^{\mathrm{Ref}}$, and the sampling model $\pi_{\theta}^{\mathrm{Old}}$. After every $\mu$ policy updates on a batch of sampled graphs $G$, the sampling model $\pi_{\theta}^{\mathrm{Old}}$ is synchronized with the latest parameters of $\pi_\theta$ to ensure stable exploration.
In contrast, T-GRPO adopts a subtree-level optimization paradigm. Rather than updating the policy at fixed intervals, it postpones gradient updates until a complete traversal and evaluation of a sampled subtree is performed. By computing gradients based on coherent and trajectory-consistent feedback, this design produces richer learning signals with enhanced contextual awareness. Once $\pi_\theta$ completes learning from the sampling results across all subtrees, we set $\pi_\theta^{\text{Old}} \gets \pi_\theta$. The complete training procedure is presented formally in Algorithm~\ref{alg:Train-TGRPO}.

\begin{algorithm}
\caption{Single sample in Tree-Group Relative Policy Optimization}
\label{alg:Train-TGRPO}
\begin{algorithmic}[1]
\Require $\pi_{\theta}^\mathrm{SFT}$, $buf\!fer$, $I, \mu$,$E_{init},S_{init},Q$
\Ensure $\pi_\theta^{*}$
\State policy model $\pi_\theta \leftarrow \pi_{\theta}^\mathrm{SFT}$
\State reference model $ \pi_{\theta}^\mathrm{Ref} \leftarrow \pi_{\theta}^\mathrm{SFT}$
\For{step = $1, \ldots, \mathrm{I}$} 
\State            Update the old policy model $\pi_{\theta}^\mathrm{Old}\leftarrow \pi_\theta$
 \State           $buf\!fer \gets \text{Searching}$ ($E_{init}$, $S_{init}$, $\emptyset$ , $Q$)
  \For{\{$Group$, $Reward$\} in $buf\!fer$}         
 \State            Compute token-level group relative advantage estimation based on $Group$ and $Reward$
 \For{GRPO iteration $=1, \ldots, \mu$} 
  \State Update $\pi_\theta$ by maximizing GRPO objective 
 \EndFor
 \EndFor
\EndFor
\end{algorithmic}
\end{algorithm}

\subsection{Experimental Settings}
\noindent{\textbf{Datasets.}} 
We evaluate our proposed MRE framework on two challenging and complementary TKGQA benchmarks. The first, \textsc{CRONQUESTIONS}~\cite{cronqa}, is a large-scale dataset constructed from Wikidata, featuring 410K questions that involve 1- to 3-hop temporal reasoning. With rich annotations of fine-grained timestamps and a diverse mix of entity- and time-centric queries, it has become a standard benchmark for temporal QA.The second, \textsc{TimeQuestions}~\cite{TwiRGCN}, unifies 13.5K questions from five existing datasets into a comprehensive benchmark that emphasizes multi-hop temporal reasoning. By covering explicit, implicit, comparative, and ordinal question types, thereby offering a rigorous testbed for evaluating models' temporal commonsense and ordinal reasoning capabilities.

\noindent{\textbf{Evaluation Metrics.}}
We adopt the commonly used evaluation metrics \textbf{Hits@1} and \textbf{Hits@10}, which are defined as follows for $K \in \{1, 10\}$:
\begin{equation}
\text{Hits@}K = \frac{1}{|\mathcal{T}|} \sum_{q \in \mathcal{T}} \mathbf{1} \left( \operatorname{rank}(q) \leq K \right),
\end{equation}
Where $\mathcal{T}$ denotes the test set. For a given question $q$, $\operatorname{rank}(q)$ is the rank assigned by the model to the correct answer within the list of candidates. The indicator function $\mathbf{1}(\cdot)$ returns 1 if the condition inside holds, and 0 otherwise.

\noindent{\textbf{Baseline Methods.}}
For the \textbf{CRONQUESTIONS} dataset, we compare MRE with several baselines, including EaE \cite{EaE}, EmbedKGQA \cite{EmbeddedKGQA}, CronKGQA \cite{cronqa}, EntityQR \cite{tempoqa}, TMA \cite{TMA}, TSQA \cite{TSQA}, CTRN\cite{ctrn}, TempoQR \cite{tempoqa} as well as language models BERT \cite{BERT}, RoBERTa \cite{roberta}, and ChatGPT. For the \textbf{TimeQuestions} dataset, the baselines comprise CronKGQA, TempoQR and TwiRGCN \cite{TwiRGCN} .
\begin{table*}[ht]
\centering
    \resizebox{\textwidth}{!}{
    \begin{tabular}{l|c|c|c|c|c|c|c|c|c|c}  
    \toprule
    \multirow{3}{*}{\textbf{Model}} & \multicolumn{5}{c|}{\textbf{Hits@1}}&\multicolumn{5}{c}{\textbf{Hits@10}} \\ 
    \cline{2-11}
    & \multirow{2}{*}{\textbf{Overall}} & \multicolumn{2}{c|}{\textbf{Question Type}} & \multicolumn{2}{c|}{\textbf{Answer Type}} & \multirow{2}{*}{\textbf{Overall}} &\multicolumn{2}{c|}{\textbf{Question Type}} & \multicolumn{2}{c}{\textbf{Answer Type}} \\  
    \cline{3-6} 
    \cline{8-11}
    & & \textbf{Complex} & \textbf{Simple} & \textbf{Entity} & \textbf{Time} &  &\textbf{Complex} & \textbf{Simple} & \textbf{Entity} & \textbf{Time} \\
    \cline{1-11}
    EmbedKGQA & 0.288 & 0.286 & 0.290 & 0.411 & 0.057 & 0.672 & 0.632 & 0.725 & 0.850 & 0.341\\
    % T-EaE-add & 0.278 & 0.257 & 0.306 & 0.313 & 0.213 & 0.663 & 0.614 & 0.729 & 0.662 & 0.665\\
    EaE & 0.288 & 0.257 & 0.329 & 0.318 & 0.231 & 0.678 & 0.623 & 0.753 & 0.668 & 0.698\\
    \cline{1-11}
    CronKGQA & 0.647 & 0.392 & 0.987 & 0.699 & 0.549 & 0.884 & 0.802 & 0.990 & 0.898 & 0.857\\
    EntityQR & 0.745 & 0.562 & 0.990 & 0.831 & 0.585 & 0.944 & 0.906 & 0.993 & 0.962 & 0.910\\
    TMA & 0.784 & 0.632 & 0.987 & 0.792 & 0.743 & 0.943 & 0.904 & 0.995 & 0.947 & 0.936\\
    TSQA & 0.831 & 0.713 & 0.987 & 0.829 & 0.836 & \underline{0.980} & \underline{0.968} & \underline{0.997} & \underline{0.981} & \underline{0.978}\\
    TempoQR & \underline{0.918} & \underline{0.864} & \underline{0.990} & \underline{0.926} & \underline{0.903} & 0.978 & 0.967 & 0.993 & 0.980 & 0.974\\
    \cline{1-11}
    BERT \textit{w/o tkg} & 0.071 & 0.086 & 0.052 & 0.077 & 0.06 & 0.213 & 0.205 & 0.225 & 0.192 & 0.253\\
    RoBERTa \textit{w/o tkg}& 0.07 & 0.086 & 0.05 & 0.082 & 0.048 & 0.202 & 0.192 & 0.215 & 0.186 & 0.231\\
    % KnowBERT & 0.226 & 0.220 & 0.238 & 0.252 & 0.177 & 0.586 & 0.539 & 0.646 & 0.582 & 0.592\\
    % ChatGPT (4-shot)  & 0.288 & 0.286 & 0.290 & 0.411 & 0.057 & 0.672 & 0.632 & 0.725 & 0.850 & 0.341\\
    ChatGPT \textit{w/o tkg} & 0.151  & 0.144 & 0.160 & 0.134 &0.182  & 0.308 & 0.308 & 0.307 & 0.257 & 0.402\\
    \cline{1-11}
    BERT \textit{w/ tkg} & 0.243 & 0.239 & 0.249 & 0.277 & 0.179 & 0.620 & 0.598 & 0.649 & 0.628 & 0.604\\
    RoBERTa \textit{w/ tkg} & 0.225 & 0.217 & 0.237 & 0.251 & 0.177 & 0.585 & 0.542 & 0.644 & 0.583 & 0.591\\
    ChatGPT \textit{w/ tkg} & 0.754  & 0.579 & 0.987 & 0.689 &0.873&0.852  & 0.746 & 0.992 & 0.808 & 0.933\\
    \cline{1-11}
    \rowcolor{gray!20}
    MRE(Ours) & \textbf{0.982}&\textbf{0.970}&\textbf{0.999}&\textbf{0.982}&\textbf{0.994}&\textbf{0.996}&\textbf{0.994}&\textbf{0.998}&\textbf{0.996}&\textbf{0.998}\\
    \bottomrule
    \end{tabular}}
    \caption{Performance comparison of different models on CRONQUESTIONS. The best and second best results are marked in \textbf{bold} and \underline{underlined}, respectively.
    \textit{w/o tkg} indicates that LMs answer the questions directly without using TKG information, and \textit{w/ tkg} indicates that LMs answer the questions with TKG background knowledge.}
\label{tab:main result}
\end{table*}

\section{Experimental Results and Analysis}
\label{sec:experiments}

\subsection{Overall Performance}
\label{subsec:overall}
\noindent\textbf{Results on \textsc{CRONQUESTIONS}.}
\begin{table}
\centering
\resizebox{0.48\textwidth}{!}{
    \begin{tabular}{l|ccccc}  
    \toprule
    \textbf{Model} & \textbf{Overall} & \textbf{Explicit} & \textbf{Implicit} & \textbf{Temporal} & \textbf{Ordinal}\\
    \midrule
    CronKGQA & 0.462 & 0.466 & 0.445 & 0.511 & 0.369 \\
    TempoQR	& 0.416	& 0.465	& 0.360 & 0.400 & 0.349 \\
    % EXAQT & 0.572 & 0.568 & 0.512 & 0.642 & 0.420\\
    TwiRGCN(average) & \textbf{0.605} & \textbf{0.602} & 0.586 & 0.641 & 0.518\\
    TwiRGCN(interval) & 0.603 & 0.599 & 0.603 & \textbf{0.646} & 0.494\\
    \rowcolor{gray!20}
    MRE(Ours) & 0.594 & 0.598 & \textbf{0.631} & 0.576& \textbf{0.578}\\
    \bottomrule
    \end{tabular}}
    \caption{Hits@1 for different models on TimeQuestions.}
\label{tab:mainresult2}
\end{table}
As shown in Table~\ref{tab:main result}, our MRE framework sets a new SOTA with 98.2\% Hits@1 and 99.6\% Hits@10 on the overall test set—outperforming the previous best (TempoQR) by +6.4\% and +1.8\%, respectively. This substantial gain highlights the effectiveness of our trajectory-level temporal reasoning. MRE generalizes well across both simple and complex questions. It achieves near-perfect accuracy on Simple questions 99.9\% (Hits@1), and consistently strong performance across Entity (98.2\%) and Time (99.4\%) answer types—demonstrating robust factual and temporal grounding. On more challenging complex multi-hop questions, MRE surpasses TempoQR by +10.6\% in Hits@1, validating the advantage of global trajectory optimization over local reasoning.
Table~\ref{category} further breaks down performance by reasoning types. Our MRE framework exhibits substantial improvements on challenging temporal categories such as \textit{before/after}, \textit{first/last}, and \textit{temporal joins}, where conventional models often struggle due to temporal ambiguity, sparse supervision, and multi-hop error propagation. These gains highlight the ability of MRE to model complex temporal dependencies and reason across distant events. At the same time, it maintains near-perfect accuracy in factoid-style subsets—achieving 99.6\% Hits@1 in both \textit{Simple-Entity} and \textit{Simple-Time}—demonstrating its robustness in handling both shallow retrieval and deep temporal inference within a unified reasoning framework.

Clearly, embedding-based methods (e.g. EmbedKGQA, EaE) perform poorly on time-centric questions ($\leq 23.1\%$ Hits@1), and language models without explicit temporal input (e.g. BERT, RoBERTa, ChatGPT \textit{w/o tkg}) score even lower (7.0\%--15.1\%). Even in temporal context, ChatGPT \textit{w/ tkg} reaches only 75.4\%, falling 22.8\% short of MRE. These results underscore the necessity of structured, trajectory-aware temporal reasoning for the TKGQA task.

\begin{table}[h]
	\centering
	\resizebox{0.48\textwidth}{!}{
	\begin{tabular}{l|c c c|c c|c}
	    \hline
	    ~&\multicolumn{3}{c|}{\textbf{Complex Question}}&\multicolumn{2}{c|}{\textbf{Simple Question}}& \\
	    \cline{2-6}
       {\textbf{Category}}&\textbf{Before/}&\textbf{First/}&\textbf{Time}&\textbf{Simple}&\textbf{Simple}& {\textbf{All}} \\
        ~&\textbf{After} & \textbf{Last} & \textbf{Join} & \textbf{Entity}& \textbf{Time} & ~ \\
        \hline
        EmbedKGQA & 0.199 & 0.324 & 0.223 & 0.421 & 0.087 & 0.288 \\
        T-EaE-add& 0.256 & 0.285 & 0.175 & 0.296 & 0.321 & 0.278\\ 
        T-EaE-replace & 0.256 & 0.288 & 0.168 & 0.318 & 0.346 & 0.288\\ 
        CronKGQA&0.288 & 0.371 &0.511 & 0.988 & 0.985&0.647 \\
        TMA &0.581 &0.627&0.675&0.988&0.987&0.784 \\ 
        TSQA &0.504 &0.721 &0.799&0.988&0.987 &0.831\\
        TempoQR &0.714 &0.853 &0.978&0.988&0.987 &0.918\\
        CTRN &0.747 &0.880 &0.897&0.991&0.987 &0.920\\
        \rowcolor{gray!20}
        MRE(Ours) &\textbf{0.926} & \textbf{0.948} &\textbf{0.994} &\textbf{0.992} &\textbf{0.996} & \textbf{0.982} \\
        \hline
	\end{tabular}
	}
    \caption{Hits@1 for different question types.}
	\label{category}
\end{table}
\noindent\textbf{Results on \textsc{TimeQuestions}.}
As shown in Table~\ref{tab:mainresult2}, MRE delivers strong overall performance on the \textsc{TimeQuestions} benchmark, achieving 59.4\% Hits@1—closely matching the best result reported by TwiRGCN while clearly outperforming all prior methods in key reasoning categories. In particular, MRE establishes SOTA results on \textit{implicit} (63.1\%, +4.5\%) and \textit{ordinal} (57.8\%, +6.0\%) questions-two of the most demanding types that require abstract temporal inference beyond surface-level timestamp matching. These improvements demonstrate MRE’s ability to handle event salience, temporal abstraction, and relative ordering more effectively than existing models.
These findings reinforce the superior generalization of MRE in complex temporal QA scenarios.

\noindent\textbf{Conclusion.}
Notably, these results highlight the effectiveness of the MRE framework in improving LLM with structured subgraph retrieval and trajectory-aware temporal reasoning. MRE enables robust, interpretable multi-hop inference, setting a new performance standard for TKGQA across both synthetic and real-world benchmarks.

\subsection{Ablation Study}
\label{subsec:ablation_study}
\begin{table}[h]
\centering
\renewcommand{\arraystretch}{0.95}
\setlength{\tabcolsep}{4pt}
\resizebox{0.48\textwidth}{!}{
\begin{tabular}{>{\centering\arraybackslash}m{3.2cm}|cc|cc}
\toprule
\textbf{Model} & \multicolumn{2}{c|}{\textbf{Hits@1}} & \multicolumn{2}{c}{\textbf{Hits@10}} \\
\cline{2-5}
~ \textbf{Variant} & \textbf{Overall} & \textbf{Complex} & \textbf{Overall} & \textbf{Complex} \\
\midrule
\rowcolor{gray!20}
Full MRE & \textbf{0.982} & \textbf{0.970} & \textbf{0.996} & \textbf{0.994} \\
\midrule
\textit{w/o} T-GRPO & 0.921 & 0.892 & 0.978 & 0.965 \\
\textit{w/o} Cold Start & 0.904 & 0.871 & 0.962 & 0.948 \\
\textit{w/o} Multi-Sampling & 0.876 & 0.842 & 0.945 & 0.931 \\
\midrule
Single-Hop & 0.812 & 0.774 & 0.910 & 0.892 \\
\bottomrule
\end{tabular}
}
\caption{
Ablation results on \textsc{CRONQUESTIONS}.  
\textit{w/o} indicates removal of the corresponding module from the full MRE pipeline.
}
\label{tab:ablation_study}
\end{table}
To evaluate the contribution of each component within the MRE framework, we perform a series of ablation studies systematically by removing or altering key modules, as summarized in Table~\ref{tab:ablation_study}. The single-hop Experiment yields only 81.2\% Hits@1 on overall questions and 77.4\% on complex ones, indicating that shallow reasoning over limited evidence chains is insufficient to handle the relational and temporal complexity inherent in TKGQA. When the T-GRPO module is removed, performance significantly drops to 92.1\% (Overall) and 89.2\% (Complex), despite the model still being trained on positive trajectories. This suggests that without structured trajectory exploration, the model tends to overfit local patterns and struggles to make globally coherent decisions. The tree-structured exploration enabled by T-GRPO, coupled with backward credit assignment, proves crucial for contrastive optimization and effective reasoning across multiple hops. Similarly, removing cold-start fine-tuning leads to a further decline in performance, underscoring the importance of supervised initialization in providing inductive bias and stabilizing early policy learning. 
Finally, disabling multi-trajectory sampling leads to the most significant performance drop, indicating that exposing the model to diverse and consistent reasoning paths is crucial for effective multi-hop temporal reasoning.
Rather than relying on a single trajectory, multi-trajectory sampling improves generalization, reduces overfitting, and enhances robustness to temporal ambiguity.
In general, the ablation results validate the complementary roles of each module and highlight their necessity to achieve strong performance in deep temporal reasoning tasks.

\subsection{Multi-hop Temporal Reasoning Depth Analysis}
\label{subsec:depth}
\begin{table}[h]
\centering
\renewcommand{\arraystretch}{0.95}
\setlength{\tabcolsep}{4pt}
\scriptsize % 字体进一步缩小
\resizebox{\columnwidth}{!}{ % 缩小整体表格宽度
\begin{tabular}{l|c|c|c}
\toprule
\textbf{Model} & \textbf{Hits@1@1-hop} & \textbf{Hits@1@2-hop} & \textbf{Hits@1@3-hop} \\
\midrule
CronKGQA & 0.991 & 0.873 & 0.512 \\
TempoQR & 0.992 & 0.952 & 0.754 \\
\rowcolor{gray!20}
\textbf{MRE (Ours)} & \textbf{0.999} & \textbf{0.981} & \textbf{0.943} \\
\bottomrule
\end{tabular}}
\caption{
Performance comparison of different models at various reasoning depths.
}
\label{tab:depth}
\end{table}
To rigorously validate the superiority of our MRE framework in multi-hop temporal reasoning, we analyze its performance across different reasoning depths on the CRONQUESTIONS benchmark:
As shown in Table~\ref{tab:depth}, MRE consistently outperforms strong baselines across all reasoning depths. It achieves near-perfect accuracy on 1-hop questions (99.9\%), and maintains robust performance on 3-hop questions (94.3\%), reducing the 3-hop error rate by over 75\% compared to TempoQR.
In particular, as the depth of reasoning increases, the performance gap between MRE and existing methods widens significantly, highlighting MRE’s superior capability in handling deep and complex multi-hop temporal queries. This advantage stems from two key design choices: 
(1) the tree-structured sampling strategy, which facilitates more effective exploration during training, 
and (2) the GRPO module, which propagates reward signals backward and optimizes each intermediate decision step to ensure global trajectory optimality. 
\subsection{RLHF Training Analysis}
\label{subsec:efficiency}
\begin{table}[h]
\centering
\resizebox{\columnwidth}{!}{
\begin{tabular}{l|c|c|c}
\toprule
\textbf{Method} & \textbf{Hits@1@10k} & \textbf{Hits@1@50k} & \textbf{Peak} \\
\midrule
PPO & 0.827 & 0.901 & 0.922 \\
GRPO (Flat) & 0.845 & 0.932 & 0.951 \\
\rowcolor{gray!20}
\textbf{MRE (Ours)} & \textbf{0.902} & \textbf{0.968} & \textbf{0.982} \\
\bottomrule
\end{tabular}
}
\caption{Training results compared with different RLHF approaches.}
\label{tab:efficiency}
\end{table}
To quantitatively assess the advantages of tree-structured reward propagation in multi-hop temporal reasoning, we compare the performances of PPO, GRPO, and T-GRPO. As reported in Table~\ref{tab:efficiency}, T-GRPO achieves a Hits@1 of 90.2\% with only 10k samples, surpassing PPO and GRPO (Flat) by 7.5\% and 5.7\%, respectively. This highlights a key limitation of GRPO (Flat). Although GRPO (Flat) replaces PPO’s advantage estimate with group-relative optimization to enable deeper exploration of the trajectory under the same training budget, it still relies on sparse reward supervision that is assigned only in the final step. 
In contrast, T-GRPO leverages a tree-structured optimization framework that propagates supervised signals along the entire reasoning path, facilitating fine-grained intermediate rewards and thereby enabling more informed policy updates. With extended training, T-GRPO further attains a peak Hits@1 of 98.2\%, demonstrating both superior sample efficiency and stronger convergence. These results confirm the effectiveness of tree-based credit assignment in guiding global policy optimization for complex multi-hop reasoning.

% \section{Conclusion}
% We propose the MRE framework to enhance multi-hop reasoning in TKGQA by integrating trajectory sampling, supervised fine-tuning, and a novel T-GRPO optimization algorithm. MRE significantly boosts the step-wise reasoning capabilities of LLMs, enabling the identification of globally optimal reasoning trajectories over TKGs. Extensive experiments across multiple benchmarks demonstrate that MRE consistently surpasses previous SOTA methods in accuracy, robustness, and interpretability, particularly on complex multi-hop queries. These results underscore the strong potential of combining preference-based optimization with structured multi-hop reasoning, paving the way for more accurate and explainable TKGQA.
% We propose the MRE framework to enhance multi-hop reasoning in TKGQA by combining trajectory sampling, supervised fine-tuning, and the T-GRPO optimization algorithm. MRE improves the ability of LLMs to identify globally optimal reasoning paths and consistently outperforms prior SOTA methods in accuracy and robustness, especially on complex multi-hop questions. These results underscore the strong potential of combining preference-based optimization with structured multi-hop reasoning, paving the way for more accurate and explainable temporal reasoning.
\section{Conclusion}
We propose MRE, a unified framework for enhancing multi-hop reasoning in TKGQA, which integrates trajectory sampling, supervised fine-tuning, and a novel T-GRPO algorithm. By jointly modeling forward exploration and backward evaluation, MRE substantially strengthens the step-wise reasoning capability of large language models, enabling the identification of globally optimal reasoning trajectories over temporal knowledge graphs. Extensive experiments across multiple benchmarks demonstrate that MRE consistently outperforms previous SOTA methods. Overall, this work highlights the effectiveness of combining preference-based optimization with structured multi-hop reasoning, and points toward a promising direction for building more accurate and explainable TKGQA systems.

\bibliography{aaai2026}

\end{document}